\documentclass[conference]{IEEEtran}
\IEEEoverridecommandlockouts
\usepackage{cite}
\usepackage{amsmath,amssymb,amsfonts}
\usepackage{algorithmic}
\usepackage{graphicx}
\usepackage{textcomp}
\usepackage{xcolor}
\def\BibTeX{{\rm B\kern-.05em{\sc i\kern-.025em b}\kern-.08em
    T\kern-.1667em\lower.7ex\hbox{E}\kern-.125emX}}
    
\usepackage{array}
\usepackage[caption=false,font=normalsize,labelfont=sf,textfont=sf]{subfig}
\usepackage{stfloats}
\usepackage{url}
\usepackage{verbatim}
\usepackage{balance}
\definecolor{deepgreen}{RGB}{0, 150, 0}
\definecolor{deepred}{RGB}{215, 0, 0}
\usepackage{algorithm}
\usepackage{arydshln}
\usepackage{booktabs}
\usepackage{subcaption}
\usepackage{multirow}
\usepackage{hyperref}
\usepackage{xcolor}

\begin{document}

\title{End-to-End Argument Mining through Autoregressive Argumentative Structure Prediction
\thanks{This paper has been accepted for publication at the IEEE/INNS International Joint Conference on Neural Networks (IJCNN) 2025. © 2025 IEEE. Personal use of this material is permitted. Permission from IEEE must be obtained for all other uses.}}

\author{
\IEEEauthorblockN{Nilmadhab Das, Vishal Vaibhav\IEEEauthorrefmark{1}, Yash Sunil Choudhary, V. Vijaya Saradhi, Ashish Anand}
\IEEEauthorblockA{Applied Machine Learning (AMaL) Lab} 
\IEEEauthorblockA{Department of Computer Science and Engineering, Indian Institute of Technology (IIT), Guwahati, India}
\IEEEauthorblockA{North Eastern Regional Institute of Science and Technology (NERIST), Arunachal Pradesh, India\IEEEauthorrefmark{1}}
\IEEEauthorblockA{\{nilmadhabdas, y.choudhary, saradhi, anand.ashish\}@iitg.ac.in, 424115@nerist.ac.in\IEEEauthorrefmark{1}}
\thanks{\IEEEauthorrefmark{1}The author carried out this research during an internship at IIT Guwahati.}
}


\maketitle

\begin{abstract}
Argument Mining (AM) helps in automating the extraction of complex argumentative structures such as Argument Components (ACs) like \textit{Premise, Claim} etc. and Argumentative Relations (ARs) like \textit{Support, Attack} etc. in an argumentative text. 
Due to the inherent complexity of reasoning involved with this task, modelling dependencies between ACs and ARs is challenging.
Most of the recent approaches formulate this task through a generative paradigm by flattening the argumentative structures. In contrast to that, this study jointly formulates the key tasks of AM in an end-to-end fashion using \textbf{Autoregressive Argumentative Structure Prediction (AASP)} framework. The proposed AASP framework is based on the autoregressive structure prediction framework that has given good performance for several NLP tasks. \textit{AASP} framework models the argumentative structures as constrained pre-defined sets of actions with the help of a conditional pre-trained language model. These actions build the argumentative structures step-by-step in an autoregressive manner to capture the flow of argumentative reasoning in an efficient way. Extensive experiments conducted on three standard AM benchmarks demonstrate that AASP achieves state-of-the-art (SoTA) results 
across all AM tasks in two benchmarks and delivers strong results in one benchmark.

\end{abstract}


\section{Introduction}

Argument mining (AM) focuses on extracting argumentative structures from discourse dynamics. 
It has applications in automated essay scoring \cite{ijcai2018p574}, legal decision support \cite{walker-etal-2018-evidence}, healthcare \cite{Mayer2020TransformerBasedAM}, etc. Early studies modelled AM as dependency parsing \cite{eger_neural_2017, ye-teufel-2021-end}. This approach involved intricate post-processing steps to match gold dependency graphs. Later in \cite{Morio2022EndtoendAM}, the authors proposed single-task and multi-task settings with a \textit{biaffine-attention} module \cite{dozat-manning-2018-simpler} to model relational structures efficiently. However, comparing each AC with every other AC led to class imbalance issues due to the majority of mappings being unrelated. Generative methods like \cite{he_generative_nodate} used flattened sequences but struggled to handle relational dependencies efficiently. In \cite{kawarada-etal-2024-argument}, the authors reformulated AM through a \textit{Structured Prediction} task-adaptation to perform text-to-text generation using \textit{Augmented Natural Language (ANL)} \cite{Paolini2021StructuredPA}. However, this approach required repeated AC spans, making it less efficient for complex structures with many relations.

Modelling argumentative relations is more complex than handling components, as structural representations vary across domains. Some datasets use tree structures for ARs \cite{stab_parsing_2017}, while others use non-tree-structured representations \cite{niculae_argument_2017}. In \cite{liu-etal-2023-argument}, the authors noted that argumentative structures resemble answers to \textit{why} questions, forming reasoning chains \textit{(a series of premises/claims)} leading to conclusions. Sequence-like approaches \cite{he_generative_nodate} or flattened-text approaches \cite{kawarada-etal-2024-argument} often fail to capture these complexities. This limitation affects performance in modeling relational structures when tackling all AM tasks jointly.

There are four key tasks of AM: \textit{(i) Argument Component Identification (\textbf{ACI})}, identifying argumentative text spans; \textit{(ii) Argument Component Classification (\textbf{ACC})}, categorizing spans as claims or premises; \textit{(iii) Argumentative Relation Identification (\textbf{ARI})}, detecting links between spans; and \textit{(iv) Argumentative Relation Classification (\textbf{ARC})}, determining the type of these links (e.g., support or attack). Prior work approached these tasks independently \cite{morio-etal-2020-towards} or partially \cite{bao-etal-2021-neural}. Recently, \textit{``End-to-End"} systems emerged to solve all tasks jointly \cite{sun-etal-2024-discourse}.


Recently, the \textit{Autoregressive Structured Prediction (ASP) framework} \cite{liu_autoregressive_2022} has achieved strong performance in several structured prediction tasks like NER, co-reference resolution, and relation extraction. It does so through a constrained set of structure-building actions with a conditional language model. The ASP framework has been tested on foundational NLP tasks only. This framework, though achieved convincing performance, has not been applied in complex reasoning tasks such as AM.

ASP framework is applicable in AM tasks as the standard datasets of AM are having tree or graph
structures. We note the following differences between foundational structured prediction tasks and AM tasks: \textbf{(I)} The target output spans in these foundational tasks are relatively shorter. In contrast, AM tasks involve longer spans, where each AC often spans across extended sections of text. \textbf{(II)} In relation extraction or co-reference resolution, the distance between two
related tokens (or entities) is typically smaller. On the other hand, AM tasks often involve related arguments that can span large distances. For example, one argument may be located at the beginning of a paragraph, while the other is at the end, with multiple intermediate ACs. These distinctions make it a meaningful and promising avenue to investigate the capability of the ASP framework in addressing AM tasks.

In this paper, we propose to use the ASP framework in solving all four key AM tasks and demonstrate its success for the \textbf{\textit{End-to-End}} AM framework. We name it as \textit{\textbf{Autoregressive Argumentative Structure Prediction (AASP)}}. To accommodate longer spans and distantly related ACs, we modify the feed-forward networks of the original ASP framework to a larger size. This modification is the key component in adaption the ASP framework to handle varying argumentative relational structures present in different AM corpora. Through AASP, we exploit the autoregressive phenomenon of a conditional language model for the AM task-specific constrained set of actions, including \textit{(i) Span-Identifying Actions}, which involve the identification of AC spans; \textit{(ii) Boundary-Pairing Actions}, which finalize the span boundaries; and \textit{(iii) Type-Labeling Actions}, which classify AC types, connect related ACs, and determine their AR types.

Training a conditional language model to predict these structure-building actions exposes the argumentative structure such a way that it allows the model to capture the dependencies within the ACs better. This explicit modelling contributes to the learning of intra-structure relationships between ACs by utilizing the semantic and contextual knowledge of the language model. Experimenting with three standard structurally distinct AM benchmarks, AASP achieves SoTA results in two benchmarks and produces higher Micro-F1 scores on relational AM tasks in the rest of the benchmark. In summary, the key contributions of this paper are:

\begin{figure*}
  \centering
  \includegraphics[width=0.85\textwidth]{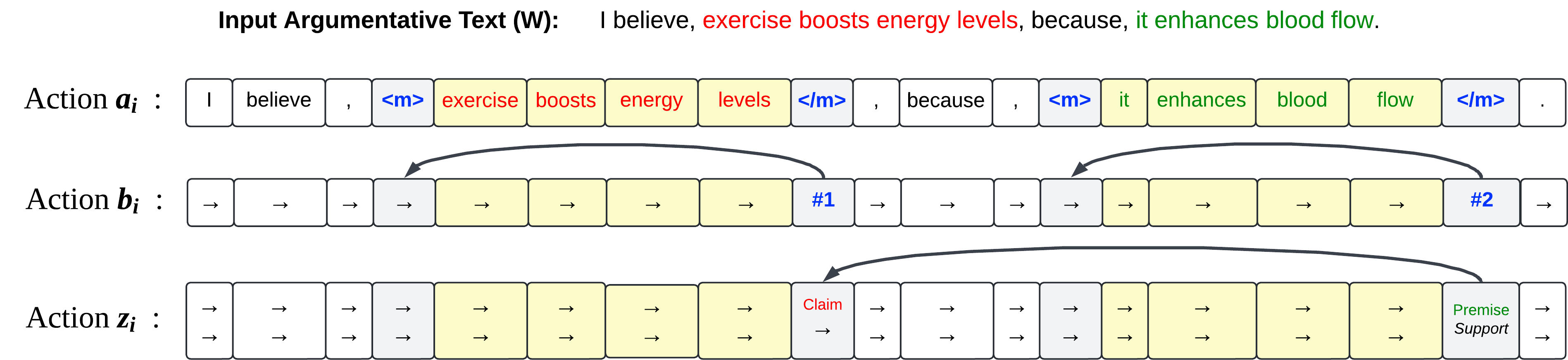}
    \caption{Illustration of the decoding process in the AASP framework for end-to-end argument mining. The process involves three main actions step-by-step: (1) Action $a_i$ constructs the target sequence for span-identifying actions; (2) Action $b_i$ finalizes the AC span boundaries by matching the \texttt{<m>} and \texttt{</m>} tokens; and (3) Action $z_i$ connects related ACs by linking their \texttt{$</m>$} tokens and classifies both the AC types and the types of their ARs.}
  \label{main-task}
\end{figure*}


\begin{enumerate}
    \item We propose \textit{Autoregressive Argumentative Structure Prediction} \textbf{(AASP)}, an end-to-end framework that jointly solves all four key tasks of AM using a conditional language model.
    \item By adapting the ASP framework for the AM tasks, we solve the ACI task using \textit{Span-Identifying} and \textit{Boundary-Pairing} actions, and the ACC, ARI and ARC tasks using the \textit{Type-Labeling} actions.
    \item Utilizing larger FFNs to execute all these actions enhances the ability to handle longer AC spans and distantly related ACs, leading to performance improvements, as demonstrated in the \textit{Ablation Study}.
    \item We conduct extensive experiments across three diverse AM benchmarks, demonstrating the robust performance of AASP in all four key AM tasks. Our results further highlight that AASP  handles relational dependencies better, thereby capturing the \textit{chain of reasoning} more effectively as a byproduct.
\end{enumerate}

\section{Related Work}
While most previous studies have concentrated on the single \cite{kuribayashi-etal-2019-empirical} or the subset of the four key tasks of AM in a pipelined setup \cite{bao_neural_2021}, recent efforts have shifted toward joint modelling of these tasks in an end-to-end manner. Among the earlier joint approaches, \textit{Persing et al.} \cite{persing_end--end_2016} first employed joint modelling of ACC and ARC tasks using Integer Linear Programming (ILP) for joint inference to mitigate error propagation between tasks. Later, \textit{Eger et al.} \cite{eger_neural_2017} explored various joint reformulations of ACC and ARC tasks using LSTM architecture, including sequence tagging, dependency parsing, multi-task tagging, and relation extraction. With the advancement of the NLP landscape, \textit{Ye et al.} \cite{ye-teufel-2021-end} first proposed a biaffine dependency parsing-based approach coupled with the rich representations of pre-trained language models to solve ACC and ARC tasks jointly. A significant limitation in the AM literature is the scarcity of datasets, which was highlighted in \cite{Morio2022EndtoendAM}, where the authors proposed a cross-corpora multi-task approach using a span-biaffine architecture with Longformer\cite{Beltagy2020LongformerTL} to first solve all four key tasks in an end-to-end setup. In this formulation, a span classifier generates BIO tags for spans, and average pooling is used to create span representations. Later, several generative frameworks also emerged to model argumentative structures as a generation of predefined targets. Within this paradigm, \textit{Bao et al.} \cite{he_generative_nodate} framed the AM task as a \textit{text-to-sequence} generation problem, producing a structured sequence comprising AC and AR types along with the start and end indices of AC spans. Later, the TANL framework \cite{Paolini2021StructuredPA} was recently adopted by \textit{Kawarada et al.} \cite{kawarada-etal-2024-argument}, framing end-to-end AM as a text-to-text generation task to model ACC and ARC tasks jointly. More recently, \textit{Sun et al.} \cite{sun-etal-2024-discourse} also proposed a generative approach with discourse structure-aware prefix using BART \cite{lewis-etal-2020-bart} to model all four key tasks with task-specific prompts.

\section{Proposed Method}

We define the argumentative paragraph as, \(W = w_1, w_2, w_3, \ldots, w_n\), where \(n\) is the total number of tokens in \(W\). We denote a text span from \(w_i\) to \(w_j\) in \(W\) as \(w_{i:j}\). Among the four key AM tasks, \textit{ACI} aims to identify the set of argumentative spans, \(C_{ACI} = \{(s_i, e_i)\}_{i=1}^{n}\), where each span \(w_{s_i:e_i}\) corresponds to a potential AC. Second, \textit{ACC} assigns a type \(t_i\) (e.g., Premise, Claim) to each identified span \((s_i, e_i) \in C_{ACI}\), resulting in \(C_{ACC} = \{(t_i, s_i, e_i)\}_{i=1}^{n}\). Third, \textit{ARI} detects the set of relations \(R_{ARI} = \{(h_j, t_j)\}_{j=1}^{m}\), where \((h_j, t_j)\) represents a connection between two ACs, with \(h_j\) denoting the head AC (the source of the relation) and \(t_j\) denoting the tail AC (the target of the relation), where \(h_j, t_j \in C_{ACC}\). Finally, \textit{ARC} categorizes each identified relation \((h_j, t_j) \in R_{ARI}\) into a specific type \(r_j\) (e.g., Support, Attack), yielding \(R_{ARC} = \{(h_j, t_j, r_j)\}_{j=1}^{m}\). 

Through the \textit{AASP framework}, we reformulate all these AM tasks as the prediction of a sequence of actions, where we construct the argumentative structure step-by-step with a conditional language model as given in Fig.~\ref{main-task}. Building on the original formulation proposed in \cite{liu_autoregressive_2022}, ASP formulation is adapted upon the four key tasks of AM for \textit{(i) Construction of action sequences} and \textit{(ii) Conditional modelling of actions}. We discuss the details in subsequent sections to maintain the coherency.

\begin{figure}
  \centering
  \includegraphics[width=0.95\columnwidth]{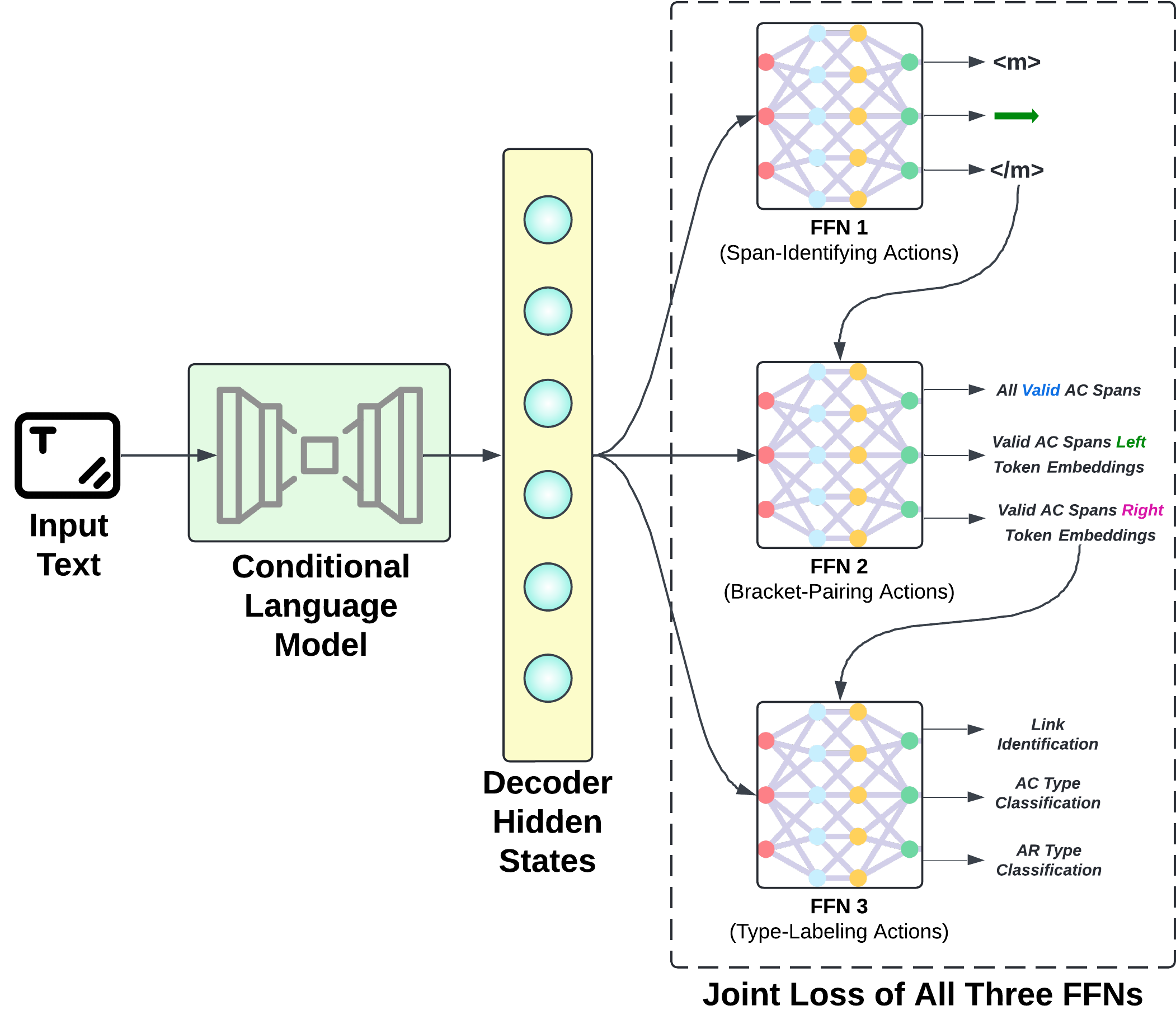}
    \caption{Model Architecture of AASP}
  \label{aasp}
\end{figure}

\subsection{Construction of Action Sequences}

Given $W$, the goal is to predict a sequence of tuples
$\langle \textit{span}_i, \textit{boundary}_i, \textit{type-label}_i \rangle$. 

Tuples denote action space. If the tuple $\langle \text{span}_i, \text{boundary}_i, \text{type-label}_i \rangle$ is denoted as $y_i$, then the goal is to predict the sequence $y_1, y_2, \dots, y_N$ and the action space $Y_i$ is given as,
\[
Y_i = A \times B_i \times Z_i \tag{1}
\]

Components of action spaces are described below.

\noindent \textbf{Span-Identifying Actions:} 

These actions take symbols from the set \( A = \{ \langle m \rangle, \rightarrow, \langle /m \rangle \} \), where:
\begin{itemize}
    \item \( \langle m \rangle \) marks the beginning of an AC span,
    \item \( \langle /m \rangle \) marks the end of the span, and
    \item \( \rightarrow \) copies the input text as it appears from left to right, irrespective of whether it is an AC or Non-AC span. 
\end{itemize}
These actions are performed using FFN 1 of Fig. \ref{aasp}.

\noindent \textbf{Boundary-Pairing Actions:} These actions denoted with set $B_n$ given as,
\[
B_n = \{ i \mid i < j \land a_i = \langle m \rangle \land a_j = \langle /m \rangle  \;\&\; i, j \in \{1,\dots,n\}\} \tag{2}
\]

Boundary-pairing actions connect the beginning of an AC span with the end of that AC span; that is, they connect the start action $\langle m \rangle$ with the end action $\langle /m \rangle$.

Specifically, the sequence produced by the \textit{Span-Identifying Actions}, is scanned from left to right, identifying each $\langle /m \rangle$ and immediately pairing it with the leftmost unmatched $\langle m \rangle$ before continuing. This way, the spans are resolved incrementally without crossing boundaries, and the unmatched symbols are removed as erroneous predictions. These actions are performed using both FFN 1 \& FFN 2 jointly, as shown in Fig. \ref{aasp}. 

With the completion of the \textit{Span-Identifying} and \textit{Boundary-Pairing} actions, the \textit{ACI task} is completed.

\noindent \textbf{Type-Labeling Actions:} These actions, denoted as \( Z_n \), perform the following: \textit{(i) labelling AC spans into predefined AC types (ACC task)}, and \textit{(ii) jointly establishing the relationships between related ACs and classifying their AR types (ARI \& ARC tasks)}. Here, the sequence produced by the \textit{Span-Identifying Actions} is traversed from left to right to find the \(\langle /m \rangle\) symbol. Once found, it connects with the previously most related already identified \(\langle /m \rangle\) to build the intra-span relationship. Additionally, both the spans corresponding to the \(\langle /m \rangle\) symbols are classified into predefined AC types, and the connected link is classified into a predefined AR type. This whole process is performed using both FFN 2 \& FFN 3 jointly in Fig. \ref{aasp}. Thus, if we define a set \( L \) that combines all AC and AR types, then:
\[
Z_n = \{ p \mid p < q \land a_p = \langle /m \rangle \} \times L \tag{3}
\]

\subsection{Conditional Modeling of Actions}
\label{conditional-modeling}
 
Given an input argumentative text \( W \), we transform it into  
a sequence of tuples denoted by \( y = (y_1, y_2, \dots, y_N) \).
This sequence of tuples is modelled as the conditional probability given below:

\[
p_\theta(y \mid W) = \prod_{n=1}^N p_\theta(y_n \mid y_{<n}, W) \tag{4}
\]
Here, each \( p_\theta(y_n \mid y_{<n}, W) \) corresponds to the probability of the \( n \)-th action given the preceding actions and the input argumentative text \( W \).

Hence, the log-likelihood of the model is then given by:
\[
\log p_\theta(y \mid W) = \sum_{n=1}^N \log p_\theta(y_n \mid y_{<n}, W) \tag{5}
\]
We then calculate the probability value of \( p_\theta(y_n \mid y_{<n}, W) \) over softmax of the dynamic set \( Y_n \) that changes as a function of the history \( y_{<n} \), i.e.,
\[
p_\theta(y_n \mid y_{<n}, W) = \frac{\exp s_\theta(y_n)}{\sum_{y'_n \in Y_n} \exp s_\theta(y'_n)} \tag{6}
\]
where \( s_\theta \) is a parameterized score function and is implemented using two independent feed-forward networks (FFN 2 \& FFN 3) as:

\[
s_\theta(y_n)
=
\begin{cases} 
    \text{FFN}_{a_n}^{z_n}(p_n) & \text{if } a_n = \langle /m \rangle \\
    \text{FFN}_{a_n}(h_n) & \text{otherwise}  \tag{7}
\end{cases}
\]
Here, \( h_n \) is the decoder hidden state at step \( n \), represented as a column vector, and \( p_n = [h_n^\top; h_{b_n}^\top]^\top \) represents the AC span mention corresponding to \( y_n \). Considering the longer text spans of ACs and the presence of distantly related ACs, we choose to select larger FFNs as compared to the ASP framework to better capture the relational dependencies.

The dynamic vocabulary is used at each time step, and the best-decoded sequence is finalized as $y^*$ using the greedy decoding strategy. At decoding step \( n \), $y_n^*$ is calculated as:
\[
y_n^* = \arg\max_{y'_n} p_\theta(y'_n \mid y_{<n}, W) \tag{8}
\]  
This resultant \( y_n^* \) is the best sequence of actions used to construct the argumentative structure for the given \( W \).

\section{Experimental Setup}

\subsection{Datasets}
We evaluate our approach on three structurally distinct standard AM benchmarks. A brief overview of these datasets is provided below.

\textbf{Argument Annotated Essay (AAE)} \cite{stab_parsing_2017}: This dataset employs a tree-structured annotation scheme where each AC can have at most one outgoing AR. It includes 402 student essays annotated at the segment (span) level, with each essay comprising multiple paragraphs. In total, there are 1,833 paragraphs, annotated with three types of ACs: \textit{Claim}, \textit{MajorClaim}, and \textit{Premise}, resulting in 6,089 ACs and 3,832 ARs.  

\textbf{Fine-Grained Argument Annotated Essay (AAE-FG)} \cite{schaefer-etal-2023-towards}: An extended version of the \texttt{AAE} dataset with finer categorization of ACs. The \textit{Claim} and \textit{MajorClaim} categories are divided into \textit{Fact}, \textit{Value}, and \textit{Policy}, while the \textit{Premise} category is split into \textit{Common Ground}, \textit{Testimony}, \textit{Hypothetical Instance}, \textit{Statistics}, \textit{Real Example}, and \textit{Others}. Hence, this version defines nine AC types, with the ARs unchanged from \texttt{AAE}.

\textbf{Consumer Debt Collection Practices (CDCP)} \cite{niculae_argument_2017}: This dataset utilizes a non-tree argumentation scheme, permitting an AC to have multiple outgoing ARs. It consists of 731 user comments sourced from the Consumer Financial Protection Bureau (CFPB) website. The data includes five types of ACs: \textit{Fact}, \textit{Testimony}, \textit{Reference}, \textit{Policy}, and \textit{Value} with two AR types: \textit{Reason} and \textit{Evidence}. In total, the dataset comprises 4,931 ACs and 1,220 ARs.  

\begin{table*}
\centering
\setlength{\tabcolsep}{12pt} 
\renewcommand{\arraystretch}{1.15}
\begin{tabular}{llccccc}
\hline
\multirow{2}{*}{\textbf{Dataset}} & \multirow{2}{*}{\textbf{Frameworks}} & \multicolumn{4}{c}{\textbf{Micro-F1 Scores of Key AM Tasks}} & \multirow{2}{*}{\textbf{AVG}} \\
\cline{3-6}
                 &                & \textbf{ACI} & \textbf{ACC} & \textbf{ARI} & \textbf{ARC} &              \\
\hline
\multirow{8}{*}{AAE} 
& BiPAM & -- & 72.90 & -- & 45.90 & -- \\
& BiPAM-syn & -- & 73.50 & -- & 46.40 & -- \\
& BART-B & 81.71 & 73.61 & 49.75 & 47.93 & 63.25 \\
& GMAM & 84.10 & 75.94 & 50.40 & 50.08 & 65.13 \\
& ST & 85.02 & 75.43 & 55.75 & 55.19 & 67.84 \\
& ANL-AM & -- & 76.93 & -- & 58.57 & -- \\
& DENIM & 85.75 & 76.50 & 59.55 & 58.51 & 70.08 \\
& \textbf{AASP (Ours)} & \textbf{87.80} & \textbf{79.29} & \textbf{63.72} & \textbf{62.69} & \textbf{73.38 (+3.3)} \\
\hline
\multirow{3}{*}{AAE-FG} 
& GMAM* & 84.20 & 57.32 & 30.81 & 25.95 & 49.57 \\
& ST* & 85.75 & 60.06 & 57.53 & 57.03 & 65.09 \\
& \textbf{AASP (Ours)} & \textbf{87.20} & \textbf{62.96} & \textbf{63.27} & \textbf{62.15} & \textbf{68.90 (+3.81)} \\
\hline
\multirow{5}{*}{CDCP} 
& BART-B & -- & 56.15 & -- & 13.76 & -- \\
& GMAM & -- & 57.72 & -- & 16.57 & -- \\
& ANL-AM & -- & \textbf{68.94} & -- & 28.42 & -- \\
& ST & \textbf{82.88} & 68.90 & 31.94 & 31.94 & \textbf{53.92} \\
& \textbf{AASP (Ours)} & 73.82 & 62.14 & \textbf{35.77} & \textbf{35.77} & 51.88 \\
\hline
\end{tabular}
\caption{Performance comparison on the \texttt{AAE}, \texttt{AAE-FG} and \texttt{CDCP} datasets. The Micro-F1 scores of ACI, ACC, ARI and ARC tasks are reported with AVG, indicating the average score of all four tasks. Apart from \texttt{AAE-FG}, all the reported results are directly taken from the corresponding baselines. For \texttt{AAE-FG}, baselines with an asterisk (*) indicate results obtained using the corresponding open-sourced code with original hyperparameters used for \texttt{AAE} dataset. ``--" indicates the results are not available for the corresponding baselines.}
\label{tab:results}
\end{table*}

\subsection{Implementation Details}
We employ the \textit{Flan-T5-Large} \cite{chung2024scaling} as the conditional language model across all experiments. For the \texttt{AAE} and \texttt{AAE-FG} datasets, we use a batch size of 64, while for \texttt{CDCP}, the batch size is set to 16. The maximum output sequence length is capped at 256 for \texttt{AAE} and \texttt{AAE-FG} and at 1024 for \texttt{CDCP}. We utilize the AdamW optimizer with a learning rate of $5 \times 10^{-5}$. All experiments are conducted on a single NVIDIA A100 GPU. Each experiment is run for 200 epochs, with checkpoints saved every 200 steps. The best model is selected based on performance on the validation set. Results are averaged over three independent runs. Both the feed-forward neural networks include one hidden layer of size 1500 and are trained with a learning rate of $3 \times 10^{-4}$. We use the following libraries: (i) ASP framework\footnote{\url{https://github.com/lyutyuh/ASP}} \cite{liu_autoregressive_2022}, and (ii) HuggingFace's Transformers\footnote{\url{https://github.com/huggingface}}.

\subsection{Evaluation}
Following the prior studies \cite{Morio2022EndtoendAM}, we evaluate all four AM tasks using the \textit{Micro-F1} score. We consider only exact matches of AC spans as correct, disregarding any partial matches to maintain strict alignment with previous works.

\subsection{Baselines}



    


    
    



We select the following SoTA models as baselines, comparing them to the datasets where applicable:
\textbf{BiPAM} \cite{ye-teufel-2021-end}: Dependency parsing for end-to-end AM using biaffine operations and BERT-base \cite{devlin-etal-2019-bert}.
\textbf{BiPAM-syn} \cite{ye-teufel-2021-end}: Extends BiPAM with syntactic information from the Stanford dependency parser.
\textbf{BART-B} \cite{yan-etal-2021-unified}: A generative model for aspect-based sentiment analysis, adapted for AM by \cite{he_generative_nodate}.
\textbf{GMAM} \cite{he_generative_nodate}: Frames AM as sequence generation with positional encoding and pointer mechanisms.
\textbf{ST} \cite{Morio2022EndtoendAM}: A dependency parser for AM using Longformer \cite{Beltagy2020LongformerTL} and biaffine attention \cite{dozat-manning-2018-simpler}.
\textbf{ANL-AM} \cite{kawarada-etal-2024-argument}: Adapts the TANL framework \cite{Paolini2021StructuredPA} for AM with text-to-text generation.
\textbf{DENIM} \cite{sun-etal-2024-discourse}: A sequence generation-based AM framework with discourse-aware multi-task prompting.

\section{Results and Discussion}

\subsection{Main Results and Comparison with Baselines}

\begin{table}
\centering
\setlength{\tabcolsep}{4pt}
\renewcommand{\arraystretch}{1.25}
\begin{tabular}{llcccccc}
\hline
\textbf{Dataset} & \textbf{Model} & \textbf{ACI} & \textbf{ACC} & \textbf{ARI} & \textbf{ARC} & \textbf{AVG} \\
\hline

\multirow{2}{*}{AAE} & AASP & \textbf{87.80} & \textbf{79.29} & \textbf{63.72} & \textbf{62.69} & \textbf{73.38} \\
 & AASP\textit{(-rhs)} & 87.61 & 78.23 & 61.49 & 60.19 & 71.88 \textcolor{deepred}{(-1.50)} \\
 \cdashline{1-7}
\multirow{2}{*}{AAE-FG} & AASP & \textbf{87.20} & 62.96 & \textbf{63.27} & \textbf{62.15} & \textbf{68.90} \\
 & AASP\textit{(-rhs)} & 86.94 & \textbf{63.43} & 62.10 & 60.78 & 68.31 \textcolor{deepred}{(-0.59)} \\
 \cdashline{1-7}
\multirow{2}{*}{CDCP} & AASP & 73.82 & 62.14 & \textbf{35.77} & \textbf{35.77} & \textbf{51.88} \\
 & AASP\textit{(-rhs)} & 73.82 & \textbf{62.85} & 33.94 & 33.94 & 51.13 \textcolor{deepred}{(-0.74)} \\

\hline
\end{tabular}
\caption{Results of ablation experiments across datasets. AASP\textit{(-rhs)} refers to the AASP framework with \textit{Reduced Hidden States} of both the Feed-Forward Networks corresponding to the score function $s_\theta$ described in Section \ref{conditional-modeling}. Best scores are in \textbf{Bold}.}
\label{table:ablation}
\end{table}

Table \ref{tab:results} compares the overall performance of AASP compared to recent baselines across all four AM tasks on the three standard datasets. Our method consistently outperforms existing baselines on the \texttt{AAE} and \texttt{AAE-FG} datasets, achieving average Micro-F1 score improvements of 3.3\% and 3.81\%, respectively. These results underscore the effectiveness of AASP in handling tree-structured argumentation. Notably, the improvements are most pronounced in the complex relational tasks (ARI and ARC), where AASP surpasses the baselines by 4.17\% and 4.18\% on \texttt{AAE}, and 5.74\% and 5.12\% on \texttt{AAE-FG}, compared to the gains in the component tasks (ACI and ACC): 2.05\% and 2.79\% on \texttt{AAE}, and 1.45\% and 2.90\% on \texttt{AAE-FG}. Specifically, in \texttt{AAE} dataset, the most recent generative SoTA baseline DENIM uses an additional discourse structure information through \textit{prefix} to better capture the relational dependency of argumentative structures. In contrast, AASP produces SoTA results without using any additional customized information. On the non-tree-structured \texttt{CDCP} dataset, AASP demonstrates significant gains in the relational tasks, achieving a substantial improvement of 3.83\% in ARI and ARC over the current SoTA. However, in component tasks such as ACI and ACC, AASP shows limited performances in the \texttt{CDCP} dataset. While most baselines struggle with complex relational tasks, AASP demonstrates improved performance in both ARI and ARC tasks in both tree-structured and non-tree-structured argumentation. This highlights AASP's ability to model relational dependencies effectively across diverse datasets.

Given that the ST baseline has been widely applied to most standard datasets, it stands as a robust baseline for comparison. Therefore, in the subsequent sections, we compare all analyses with ST to better compare the performance of AASP.

\subsection{Ablation Study}

We conduct an ablation study to assess the impact of FFNs in the score function $s_\theta$ by altering their size. In AASP, the hidden state size is 1500, while the original ASP framework uses 150. To analyze this, we run experiments on all datasets with a hidden state size of 150 instead of 1500. We denote this variant as AASP\textit{(-rhs)}, where \textit{rhs} stands for \textit{reduced hidden states}. The results, as shown in Table~\ref{table:ablation}, indicate that AASP\textit{(-rhs)} underperforms in terms of average scores across all four AM tasks compared to AASP. The ACI task performance is dropped in \texttt{AAE} and \texttt{AAE-FG} datasets with smaller FFNs. It means that the \textit{Boundary-Pairing} actions require larger FFNs to finalise the boundaries efficiently. Although for \texttt{AAE-FG} and \texttt{CDCP}, the ACC task performance improves in AASP\textit{(-rhs)}, the performance decline in ARI and ARC tasks is significant across all datasets. It suggests that the size of the FFNs plays a crucial role in effectively modelling complex relational dependencies. Specifically, the score function $s_\theta$ is better equipped to handle distantly related ACs when the \textit{Type-Labeling} actions connect the $\langle /m \rangle$ symbols of both the ACs. As a by-product, the \textit{chains of reasoning} is also handled efficiently when larger FFNs are used.

\begin{figure*}
  \centering
  \includegraphics[width=0.7\textwidth]{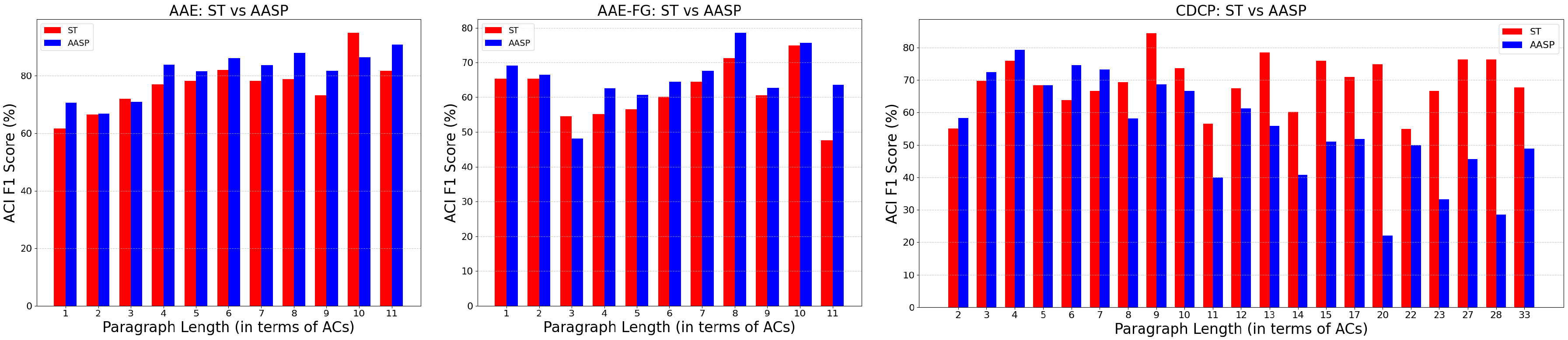}
    \caption{Comparison of ACI Micro-F1 scores (\%) between AASP and ST across paragraphs of varying lengths (in terms of the number of ACs present in that paragraph) for the \texttt{AAE}, \texttt{AAE-FG}, and \texttt{CDCP} datasets.}
  \label{length_vs_aci_f1}
\end{figure*}

\begin{figure*}
  \centering
  \includegraphics[width=0.6\textwidth]{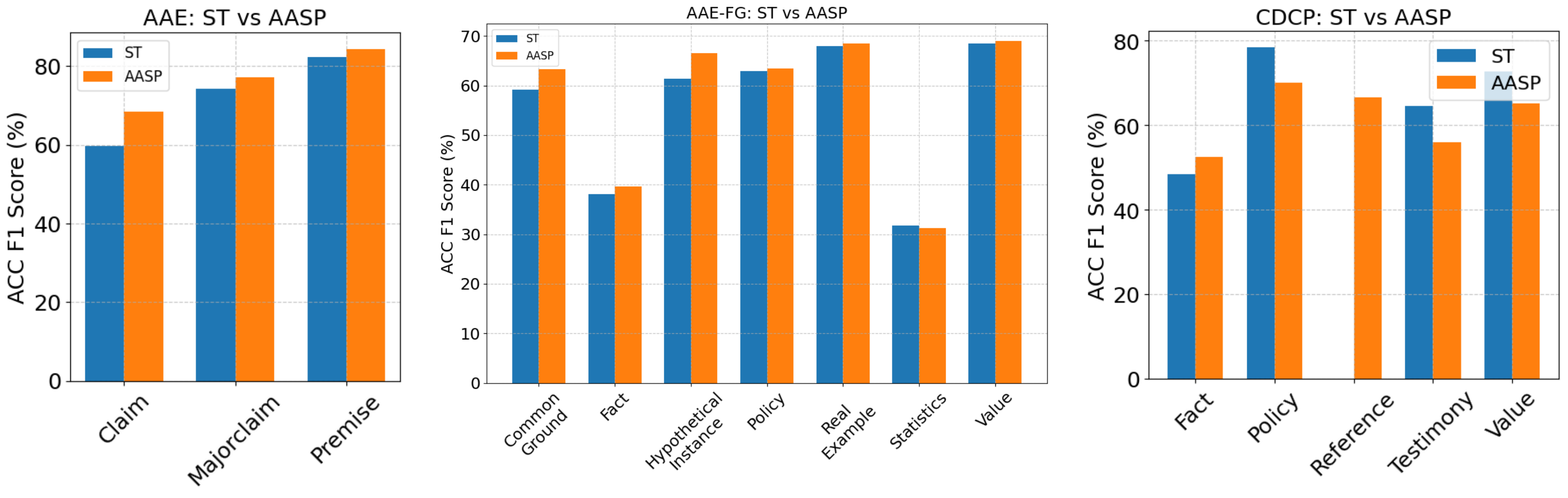}
    \caption{Comparison of ACC Micro-F1 scores (\%) for AASP and ST across different AC categories in the \texttt{AAE}, \texttt{AAE-FG}, and \texttt{CDCP} datasets.}
  \label{acc_comparison}
\end{figure*}

\begin{figure*}
  \centering
  \includegraphics[width=0.7\textwidth]{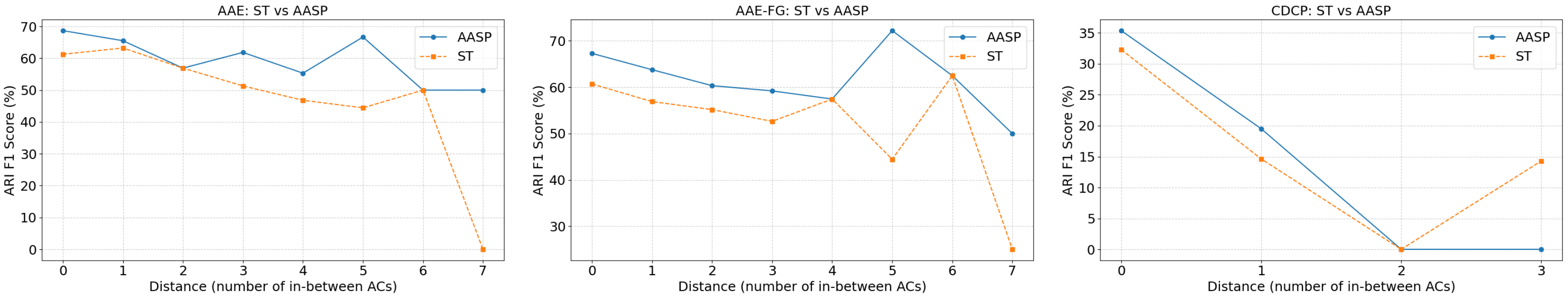}
    \caption{Performance analysis of ARI task Micro-F1 scores (\%) for AASP and ST across varying distances (in terms of number of intermediate ACs) in the AAE, AAE-FG, and CDCP datasets.}
  \label{ari_vs_distance}
\end{figure*}

\subsection{Performance of the ACI task on Varied Length Paragraphs}
In this section, we compare the performance of AASP with ST for the ACI task across datasets (See Fig.~\ref{length_vs_aci_f1}). In the \texttt{AAE} dataset, the Micro-F1 score for the ACI task improves as paragraph length (in ACs) increases, peaking at 9--11 ACs. The ST baseline starts at around 60\% Micro-F1 for single-AC paragraphs and rises to 80\% for longer ones. AASP begins slightly higher, at 65\%, and consistently outperforms ST, exceeding 80\% for the longest paragraphs. For the \texttt{AAE-FG} dataset, a similar trend is observed, but the overall performance is slightly lower than in \texttt{AAE}. Here, both ST and AASP show a more gradual improvement, beginning at around 65\% and reaching approximately 70--75\% for 8--10 ACs. In the \texttt{CDCP} dataset, the relationship between paragraph length and Micro-F1 score becomes interesting. The ST baseline performs well across a wide range of lengths, achieving high scores (70--80\%) for most configurations. However, while AASP remains competitive, it underperforms compared to ST for longer paragraphs, especially beyond 8 ACs. But it surpasses ST by a noticeable margin for paragraphs with fewer than 8 ACs. This indicates that the AASP framework struggles with the ACI task for longer paragraphs in the complex, non-tree-structured \texttt{CDCP} dataset. However, it achieves superior performance across all three datasets for paragraphs of moderate length.

\subsection{Analysis of the ACC task for Different AC Categories}

In this section, we compare the performance of AASP and ST on different AC categories across datasets (see Fig.~\ref{acc_comparison}). In the \texttt{AAE} dataset, AASP outperforms ST across all AC types. The \texttt{PREMISE} category achieves the highest Micro-F1 score, making it the most predictable for both frameworks. Strong performance is also observed for the \texttt{MAJORCLAIM} type. Notably, AASP shows a significant improvement of nearly 10\% over ST for the \texttt{CLAIM} category, making it the better choice for \texttt{CLAIM} classification. The \texttt{AAE-FG} dataset, with its more fine-grained AC labels, poses a greater challenge for both frameworks. In the \texttt{AAE-FG} dataset, the original AC categories from \texttt{AAE} are divided into finer-grained classes, reducing the training instances per class. Despite this, both frameworks perform robustly. AASP shows notable gains in categories like \texttt{HYPOTHETICAL INSTANCE} and \texttt{COMMON GROUND}, demonstrating its strength in handling finer-grained argumentation. The \texttt{CDCP} dataset presents mixed results. While ST performs better overall in the AC classification task, AASP outperforms ST in certain categories. For example, AASP achieves better performance in \texttt{FACT} classification. Surprisingly, ST fails to detect any instances of the \texttt{REFERENCE} category, whereas AASP achieves over 65\% Micro-F1 score in this category. These empirical results emphasize the effectiveness of AASP for ACC task, particularly for fine-grained or underrepresented categories.

\subsection{Analysis of ARI Task for Long-Range Relations}
In argumentative paragraphs, the relationships between related ACs exhibit a fascinating variety of patterns. Some related ACs are directly linked with no intermediate ACs, forming straightforward connections, while others are moderately far apart, involving a few intermediate ACs. The most challenging cases arise when ACs are distantly related, with one appearing near the beginning of the paragraph and the other near the end. Identifying such varied-distant relations is critical for understanding complex argumentative structures. To assess this capability, we analyze the performance of both ST and AASP across varying distantly related ACs, as shown in Fig. \ref{ari_vs_distance}. In the \texttt{AAE} dataset, AASP consistently outperforms ST across all distances. For direct relations (no intermediate ACs), AASP achieves a Micro-F1 score of nearly 70\%, while ST lags behind at 60\%. Although performance declines with increasing distance, AASP proves more robust. It maintains a Micro-F1 score above 50\% even at the maximum distance of 7 intermediate ACs, where ST experiences a sharp drop. In the \texttt{AAE-FG} dataset, a similar trend is observed. The CDCP dataset presents a more challenging scenario due to the complex, non-tree-structured nature of its argumentative relations. Both frameworks show similar patterns in this dataset up to 2 intermediate ACs. However, for the maximum distance (3 intermediate ACs), AASP’s performance drops sharply to 0\%, highlighting the challenges of identifying long-range relations in this dataset.

\begin{table*}
\centering
\begin{tabular}{llcccccc}
\toprule
\multirow{2}{*}{\textbf{Dataset}} & \multirow{2}{*}{\textbf{Framework}} & \multicolumn{3}{c}{\textbf{Lengthwise Distributions of Chains}} & \multicolumn{2}{c}{\textbf{Accuracy (\%)}} \\ 
\cmidrule(lr){3-5} \cmidrule(lr){6-7}
                 &                    & \textbf{Ground Truth} & \textbf{Predicted} & \textbf{Correct} & \textbf{2-length} & \textbf{3-length} \\ \midrule

\multirow{2}{*}{AAE} 
& ST   & \{2: 130, 3: 15, 4: 2\} & \{2: 109, 3: 5\} & \{2: 39\} & 30.00 & 0.00 \\
& AASP & \{2: 130, 3: 15, 4: 2\} & \{2: 124, 3: 16, 4: 1\} & \{2: 46, 3: 2\} & \textbf{35.38} & \textbf{13.33} \\ \midrule

\multirow{2}{*}{AAE-FG} 
& ST   & \{2: 130, 3: 15, 4: 2\} & \{2: 122, 3: 9, 4: 1\} & \{2: 37, 3: 2\} & 28.46 & 13.33 \\
& AASP & \{2: 130, 3: 15, 4: 2\} & \{2: 125, 3: 10\}       & \{2: 45, 3: 3\} & \textbf{34.62} & \textbf{20.00} \\ \midrule

\multirow{2}{*}{CDCP} 
& ST   & \{2: 23, 3: 9, 4: 1\}  & \{2: 18, 3: 10\}        & \{2: 1\}        & 4.35  & 0.00 \\
& AASP & \{2: 23, 3: 9, 4: 1\}  & \{2: 13, 3: 1\}         & \{2: 4\}        & \textbf{17.39} & 0.00 \\ \bottomrule
\end{tabular}
\caption{Performance Comparison of ST and AASP across datasets on Relation Chain Detection for Varied Length Chains.}
\label{chains}
\end{table*}

\begin{figure}
  \centering
  \includegraphics[width=0.6\columnwidth]{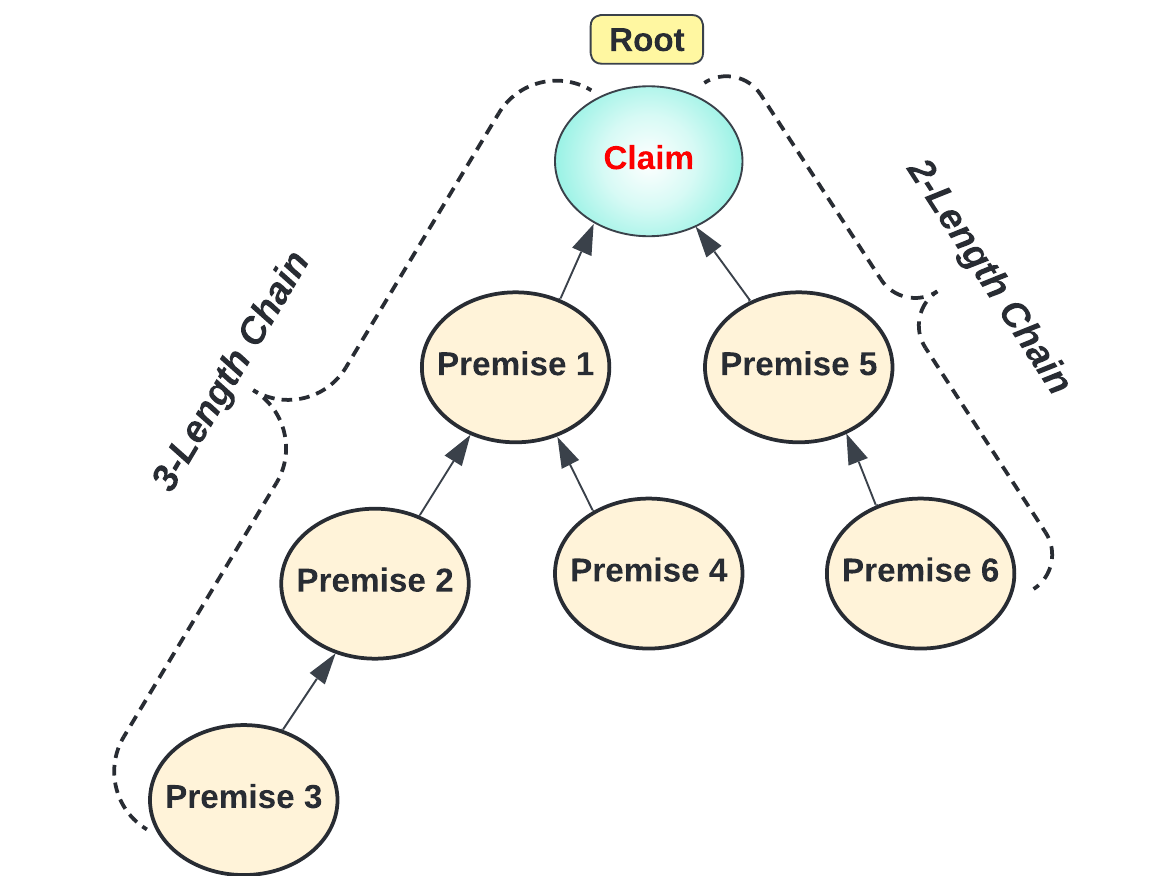}
    \caption{Example of chains present in an argumentative structure}
  \label{chain_diagram}
\end{figure}

\subsection{Performance Analysis of Relational Chains}

Relational chains (e.g. a series of premises/claims as shown in Fig.~\ref{chain_diagram}) are the building blocks of the \textit{Chain of Reasoning} present in any argumentative discourse. Currently, none of the existing works has focused on this aspect of analysis. 
As shown in Fig.~\ref{chain_diagram}, an argumentative paragraph comprising seven ACs: $\{Claim, Premise_1, Premise_2, \ldots, Premise_6\}$. With these ACs, a 3-length sequential chain is formed as: \(Premise_3 \rightarrow Premise_2 \rightarrow Premise_1 \rightarrow Claim\), with \(Claim\) as the root AC. Here, each AC is directly connected to the preceding one. In such configurations, identifying relations between adjacent ACs poses a significant challenge. This difficulty arises because all ACs in the chain are (in-)directly connected through the root AC, increasing the likelihood of misidentifying relations. For example, one might incorrectly predict \(Premise_3 \rightarrow Premise_1\) instead of the correct \(Premise_3 \rightarrow Premise_2\). In order to assess the performance in detecting these relational chains, we present an in-depth analysis in Table \ref{chains}, where we compare ST and AASP frameworks across datasets. AASP achieves 35.38\% accuracy for 2-length chains in \texttt{AAE} and 34.62\% in \texttt{AAE-FG}, compared to ST's 30.00\% and 28.46\%, respectively. However, both frameworks struggle to accurately identify longer chains (3-length and above). The CDCP dataset, with its complex non-tree argumentative structures, poses challenges in both frameworks. However, AASP still surpasses ST with 17.39\% accuracy for 2-length chains compared to ST's 4.35\%. This comparative analysis highlights the effectiveness of AASP in detecting chains of ACs over existing baselines. Moreover, it also emphasizes the need for further advancements in modelling complex relational structures to capture the chain of reasoning more accurately.

\begin{figure*}
  \centering
  \includegraphics[width=0.7\textwidth]{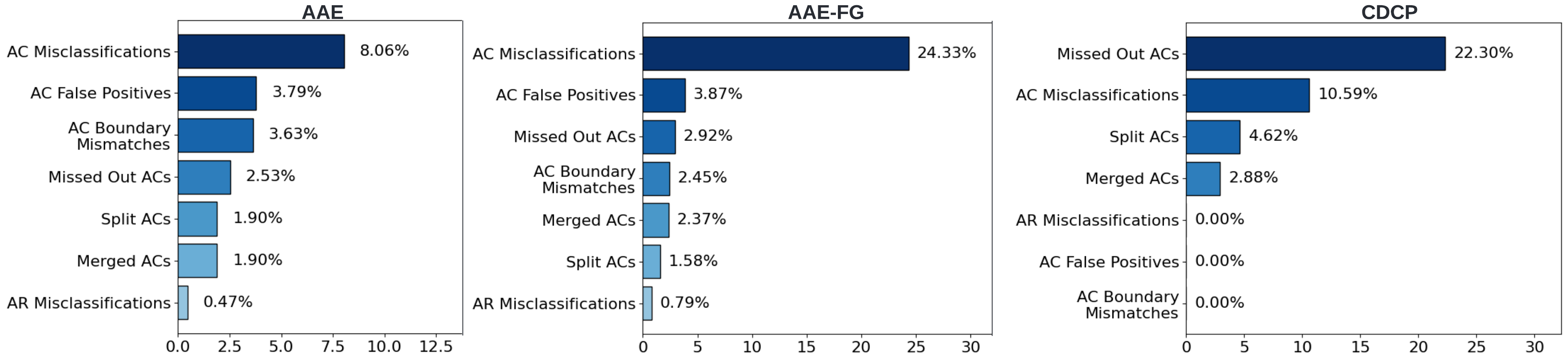}
    \caption{Error distribution across three datasets with AASP framework, highlighting the percentage of various error categories.}
  \label{errors}
\end{figure*}

\subsection{Error Analysis}

We conducted error analysis across all three datasets and identified the following errors (see Fig. \ref{errors}): \textit{(i) AC Boundary Mismatches}, where predicted ACs differ in length from the ground truth; \textit{(ii) Missed ACs}, where valid ACs are not detected; \textit{(iii) False Positive ACs}, where non-existent ACs are predicted; \textit{(iv) AC Misclassifications}, where ACs are assigned incorrect types; \textit{(v) AR Misclassifications}, where ARs are incorrectly labeled; \textit{(vi) Split ACs}, where true ACs are divided into smaller segments; and \textit{(vii) Merged ACs}, where multiple ACs are wrongly combined into one.

Analyzing the errors across the datasets, \textit{AC Misclassifications} emerge as the most significant error type overall, particularly in AAE-FG (24.33\%) and CDCP (10.59\%), indicating a consistent challenge in correctly classifying ACs, where the number of AC types is more. \textit{Missed Out ACs}, another major source of error for the CDCP dataset with 22.30\% error, suggests that the AASP struggles more with identifying all potential ACs in this complex non-tree-structured dataset than other datasets. \textit{AC False Positives} and \textit{AC Boundary Mismatches} show relatively similar proportions in AAE (3.79\% and 3.63\%, respectively) and AAE-FG (3.87\% and 2.45\%, respectively). Surprisingly, the CDCP dataset has no such errors due to the fact that in CDCP, each AC spans over a full sentence, unlike AAE or AAE-FG, where the ACs are further segmented inside sentences. Detecting full-sentence ACs is relatively easier than the segmented ACs. Upon manually analyzing the predictions of both versions of AAE, we find spans such as \textit{``I believe that"} or \textit{``People might argue that"} are included inside the AC spans, alleviating the incorrect boundaries. \textit{Split ACs} and \textit{Merged ACs}, which reflect structural prediction errors, remain minor across datasets but still noticeable, with AAE and AAE-FG showing similar patterns (1.9\%-2.4\% for both error types). However, the CDCP dataset shows a slightly higher occurrence of Split ACs (4.62\%) compared to Merged ACs (2.88\%) due to its ACs spanning over a full sentence. Whenever the sentence is longer, the split sometimes happens when connectives such as \textit{``and"} or \textit{``because"} are present inside sentences. Surprisingly, \textit{AR Misclassifications} are the least frequent errors across all datasets, staying below 1\%, indicating that the AASP performs relatively well in recognizing argumentative relations effectively.



\section{Conclusion}  
In this work, we introduced \textbf{AASP}, an end-to-end framework designed to address all four key tasks of AM jointly. By extending the ASP framework, we effectively tackled the ACI task through span identification and boundary-pairing actions, while ACC, ARI, and ARC tasks were addressed using type-labeling actions. Our approach leverages larger FFNs to enhance the execution of these actions, improving the model’s ability to process longer AC spans and distantly related ACs. This contributes to overall performance gains, as verified through our ablation study. To gain deeper insights into the relational performance, we conduct an in-depth relational chain analysis. The results highlight the effectiveness of AASP in outperforming competitive baselines. Out of the three benchmarks, AASP achieved SoTA results across all four AM tasks in two benchmarks while demonstrating SoTA performance for relational tasks in the remaining benchmark.

\bibliography{custom}

\begin{thebibliography}{10}
\providecommand{\url}[1]{#1}
\csname url@samestyle\endcsname
\providecommand{\newblock}{\relax}
\providecommand{\bibinfo}[2]{#2}
\providecommand{\BIBentrySTDinterwordspacing}{\spaceskip=0pt\relax}
\providecommand{\BIBentryALTinterwordstretchfactor}{4}
\providecommand{\BIBentryALTinterwordspacing}{\spaceskip=\fontdimen2\font plus
\BIBentryALTinterwordstretchfactor\fontdimen3\font minus
  \fontdimen4\font\relax}
\providecommand{\BIBforeignlanguage}[2]{{%
\expandafter\ifx\csname l@#1\endcsname\relax
\typeout{** WARNING: IEEEtran.bst: No hyphenation pattern has been}%
\typeout{** loaded for the language `#1'. Using the pattern for}%
\typeout{** the default language instead.}%
\else
\language=\csname l@#1\endcsname
\fi
#2}}
\providecommand{\BIBdecl}{\relax}
\BIBdecl

\bibitem{ijcai2018p574}
Z.~Ke, W.~Carlile, N.~Gurrapadi, and V.~Ng, ``Learning to give feedback:
  Modeling attributes affecting argument persuasiveness in student essays,'' in
  \emph{IJCAI}, 2018.

\bibitem{walker-etal-2018-evidence}
V.~R. Walker, D.~Foerster, J.~M. Ponce, and M.~Rosen, ``Evidence types,
  credibility factors, and patterns or soft rules for weighing conflicting
  evidence: Argument mining in the context of legal rules governing evidence
  assessment,'' in \emph{Proceedings of the 5th Workshop on Argument Mining},
  2018.

\bibitem{Mayer2020TransformerBasedAM}
T.~Mayer, E.~Cabrio, and S.~Villata, ``Transformer-based argument mining for
  healthcare applications,'' in \emph{ECAI}, 2020.

\bibitem{eger_neural_2017}
S.~Eger, J.~Daxenberger, and I.~Gurevych, ``Neural {End}-to-{End} {Learning}
  for {Computational} {Argumentation} {Mining},'' in \emph{ACL}, 2017.

\bibitem{ye-teufel-2021-end}
Y.~Ye and S.~Teufel, ``End-to-end argument mining as biaffine dependency
  parsing,'' in \emph{EACL}, 2021.

\bibitem{Morio2022EndtoendAM}
G.~Morio, H.~Ozaki, T.~Morishita, and K.~Yanai, ``End-to-end argument mining
  with cross-corpora multi-task learning,'' \emph{TACL}, 2022.

\bibitem{dozat-manning-2018-simpler}
T.~Dozat and C.~D. Manning, ``Simpler but more accurate semantic dependency
  parsing,'' in \emph{ACL}, 2018.

\bibitem{he_generative_nodate}
J.~Bao, Y.~He, Y.~Sun, B.~Liang, J.~Du, B.~Qin, M.~Yang, and R.~Xu, ``A
  {Generative} {Model} for {End}-to-{End} {Argument} {Mining} with
  {Reconstructed} {Positional} {Encoding} and {Constrained} {Pointer}
  {Mechanism},'' in \emph{EMNLP}, 2022.

\bibitem{kawarada-etal-2024-argument}
M.~Kawarada, T.~Hirao, W.~Uchida, and M.~Nagata, ``Argument mining as a
  text-to-text generation task,'' in \emph{EACL}, 2024.

\bibitem{Paolini2021StructuredPA}
G.~Paolini, B.~Athiwaratkun, J.~Krone, J.~Ma, A.~Achille, R.~Anubhai, C.~N. dos
  Santos, B.~Xiang, and S.~Soatto, ``Structured prediction as translation
  between augmented natural languages,'' \emph{ArXiv}, 2021.

\bibitem{stab_parsing_2017}
C.~Stab and I.~Gurevych, ``Parsing {Argumentation} {Structures} in {Persuasive}
  {Essays},'' \emph{Computational Linguistics}, 2017.

\bibitem{niculae_argument_2017}
V.~Niculae, J.~Park, and C.~Cardie, ``Argument {Mining} with {Structured}
  {SVMs} and {RNNs},'' in \emph{ACL}, 2017.

\bibitem{liu-etal-2023-argument}
B.~Liu, V.~Schlegel, R.~Batista-Navarro, and S.~Ananiadou, ``Argument mining as
  a multi-hop generative machine reading comprehension task,'' in \emph{EMNLP},
  2023.

\bibitem{morio-etal-2020-towards}
G.~Morio, H.~Ozaki, T.~Morishita, Y.~Koreeda, and K.~Yanai, ``Towards better
  non-tree argument mining: Proposition-level biaffine parsing with
  task-specific parameterization,'' in \emph{ACL}, 2020.

\bibitem{bao-etal-2021-neural}
J.~Bao, C.~Fan, J.~Wu, Y.~Dang, J.~Du, and R.~Xu, ``A neural transition-based
  model for argumentation mining,'' in \emph{ACL-IJCNLP}, 2021.

\bibitem{sun-etal-2024-discourse}
Y.~Sun, G.~Chen, C.~Yang, J.~Bao, B.~Liang, X.~Zeng, M.~Yang, and R.~Xu,
  ``Discourse structure-aware prefix for generation-based end-to-end
  argumentation mining,'' in \emph{ACL}, 2024.

\bibitem{liu_autoregressive_2022}
T.~Liu, Y.~Jiang, N.~Monath, R.~Cotterell, and M.~Sachan, ``Autoregressive
  structured prediction with language models,'' in \emph{EMNLP}, 2022.

\bibitem{kuribayashi-etal-2019-empirical}
T.~Kuribayashi, H.~Ouchi, N.~Inoue, P.~Reisert, T.~Miyoshi, J.~Suzuki, and
  K.~Inui, ``An empirical study of span representations in argumentation
  structure parsing,'' in \emph{ACL}, 2019.

\bibitem{bao_neural_2021}
J.~Bao, C.~Fan, J.~Wu, Y.~Dang, J.~Du, and R.~Xu, ``A {Neural}
  {Transition}-based {Model} for {Argumentation} {Mining},'' in
  \emph{ACL-IJCNLP}, 2021.

\bibitem{persing_end--end_2016}
I.~Persing and V.~Ng, ``End-to-{End} {Argumentation} {Mining} in {Student}
  {Essays},'' in \emph{NAACL-HLT}, 2016.

\bibitem{Beltagy2020LongformerTL}
I.~Beltagy, M.~E. Peters, and A.~Cohan, ``Longformer: The long-document
  transformer,'' \emph{ArXiv}, 2020.

\bibitem{lewis-etal-2020-bart}
M.~Lewis, Y.~Liu, N.~Goyal, M.~Ghazvininejad, A.~Mohamed, O.~Levy, V.~Stoyanov,
  and L.~Zettlemoyer, ``{BART}: Denoising sequence-to-sequence pre-training for
  natural language generation, translation, and comprehension,'' in \emph{ACL},
  2020.

\bibitem{schaefer-etal-2023-towards}
R.~Schaefer, R.~Knaebel, and M.~Stede, ``Towards fine-grained argumentation
  strategy analysis in persuasive essays,'' in \emph{Proceedings of the 10th
  Workshop on Argument Mining}, 2023.

\bibitem{chung2024scaling}
H.~W. Chung, L.~Hou, S.~Longpre, B.~Zoph, Y.~Tay, W.~Fedus, Y.~Li, X.~Wang,
  M.~Dehghani, S.~Brahma \emph{et~al.}, ``Scaling instruction-finetuned
  language models,'' \emph{JMLR}, 2024.

\bibitem{devlin-etal-2019-bert}
J.~Devlin, M.-W. Chang, K.~Lee, and K.~Toutanova, ``{BERT}: Pre-training of
  deep bidirectional transformers for language understanding,'' in
  \emph{NAACL-HLT}, 2019.

\bibitem{yan-etal-2021-unified}
H.~Yan, J.~Dai, T.~Ji, X.~Qiu, and Z.~Zhang, ``A unified generative framework
  for aspect-based sentiment analysis,'' in \emph{ACL-IJCNLP}, 2021.

\end{thebibliography}
\bibliographystyle{IEEEtran}

\end{document}